\documentclass[algorithms,article,moreauthors,pdftex,preprint,a4paper]{Definitions/mdpi}

\firstpage{1}
\pubvolume{00}
\issuenum{1}
\articlenumber{5}
\pubyear{2020}
\copyrightyear{2020}
\history{Received: 18 November 2020; Accepted: 7 December 2020; Published: date}

\pdfoutput=1

\usepackage{amsmath}
\usepackage{amssymb}
\usepackage[ruled,vlined]{algorithm2e}

\setitemize{parsep=6pt,itemsep=0pt,leftmargin=*,labelsep=1em}
\setenumerate{parsep=6pt,itemsep=0pt,leftmargin=*,labelsep=1em}
\setlist[description]{itemsep=0mm}

\usepackage{algpseudocode}

\newlength\hrulethickness
\makeatletter
\renewcommand\fs@ruled{\def\@fs@cfont{\bfseries}\let\@fs@capt\floatc@ruled
  \def\@fs@pre{\hrule height \hrulethickness depth0pt \kern2pt}%
  \def\@fs@post{\kern2pt\hrule height \hrulethickness depth0pt \relax}%
  \def\@fs@mid{\kern2pt\hrule height \arrayrulewidth depth0pt \kern2pt}%
  \let\@fs@iftopcapt\iftrue}
\makeatother

\newcommand{\mref}[1]{(\ref{#1})}
\newcommand{\V}{\mathbf{V}}
\newcommand{\X}{\mathbf{X}}
\renewcommand{\G}{\mathcal{G}}
\newcommand{\D}{\mathcal{D}}
\newcommand{\Prob}{\operatorname{P}}
\newcommand{\given}{\operatorname{\mid}}
\newcommand{\indep}{\perp\hspace{-0.6em}\perp}
\newcommand{\notindep}{\perp\hspace{-0.7em}\not\hspace{0.07em}\perp}
\newcommand{\Pai}{\mathit{Pa}(X_i)}
\newcommand{\T}{\Theta}
\newcommand{\hT}{\widehat{\T}}
\newcommand{\Ti}{\Theta_{X_i}}
\newcommand{\E}{\operatorname{E}}
\newcommand{\piijk}{\pi_{ik \given j}}
\newcommand{\hpiijk}{\widehat{\pi}_{ik \given j}}
\newcommand{\aijk}{\alpha_{ijk}}
\newcommand{\KL}{\operatorname{KLD}}
\newcommand{\klsep}{\operatorname{||}}

\Title{Hard and Soft EM in Bayesian Network Learning from Incomplete Data}

\Author{Andrea Ruggieri $^{1,\dagger}$, Francesco Stranieri $^{1,\dagger}$,
  Fabio Stella $^{1}$ and Marco Scutari $^{2, *}$}
\AuthorNames{Andrea Ruggieri, Francesco Stranieri, Fabio Stella, Marco Scutari}

\address{%
$^{1}$ \quad Department of Informatics, Systems and Communication,
         Universit\`a degli Studi di Milano-Bicocca, 20126 Milano MI, Italy;
         a.ruggieri4@campus.unimib.it; f.stranieri1@campus.unimib.it;
         fabio.stella@unimib.it \\
$^{2}$ \quad Istituto Dalle Molle di Studi sull'Intelligenza Artificiale
         (IDSIA), 6962 Lugano, Switzerland}
\corres{Correspondence: scutari@idsia.ch}
\firstnote{These authors contributed equally to this work.}

\abstract{
  Incomplete data are a common feature in many domains, from clinical trials to
  industrial applications. Bayesian networks (BNs) are often used in these
  domains because of their graphical and causal interpretations. BN parameter
  learning from incomplete data is usually implemented with the
  Expectation-Maximisation algorithm (EM), which computes the relevant
  sufficient statistics (``soft EM'') using belief propagation. Similarly, the
  Structural Expectation-Maximisation algorithm (Structural EM) learns the
  network structure of the BN from those sufficient statistics using algorithms
  designed for complete data. However, practical implementations of parameter
  and structure learning often impute missing data (``hard EM'') to compute
  sufficient statistics instead of using belief propagation, for both ease of
  implementation and computational speed. In this paper, we investigate the
  question: what is the impact of using imputation instead of belief propagation
  on the quality of the resulting BNs? From a simulation study using synthetic
  data and reference BNs, we find that it is possible to recommend one approach
  over the other in several scenarios based on the characteristics of the data.
  We then use this information to build a simple decision tree to guide
  practitioners in choosing the EM algorithm best suited to their problem.
}

\begin{document}

\section{Introduction}

The performance of machine learning models is highly dependent on the quality of
the data that are available to train them: the more information they contain,
the~better the insights we can obtain from them. Incomplete data contain,
by~construction, less useful information to model the phenomenon we are studying
because there are fewer complete observations from which to learn the
distribution of the variables and their interplay. Therefore, it is important to
make the best possible use of such data by incorporating incomplete observations
and the stochastic mechanism that leads to certain variables not being observed
in the~analysis.

There is ample literature on the statistical modelling of incomplete data. While
it is tempting to simply replace missing values as a separate preprocessing
step, it has long been known that even fairly sophisticated techniques like
hot-deck imputation are problematic~\cite{hotdeck}. Just deleting incomplete
samples can also bias learning, depending on how missing data are
missing~\cite{raghunathan}. Therefore, modern~probabilistic approaches have
followed the lead of Rubin~\cite{rubin1,rubin2} and modelled missing values
along with the stochastic mechanism of missingness. This class of approaches
introduces one auxiliary variable for each experimental variable that is not
completely observed in order to model the distribution of missingness; that is,
the~binary pattern of values being observed or not for that specific
experimental variable. These~auxiliary variables are then integrated out in
order to compute the expected values of the parameters of the model. The~most
common approach to learn machine learning models that build on this idea is the
\emph{Expectation-Maximisation} (EM) algorithm~\cite{em}; other approaches
such as variational inference~\cite{zoubin4} have seen substantial applications
but are beyond the scope of this paper and we will not discuss them further.
The~EM algorithm comprises two steps that are performed iteratively until
convergence: the ``expectation'' (E) step computes the expected values of the
sufficient statistics  given the current model, and~the ``Maximisation'' (M)
step updates the model with new parameter~estimates. (A~statistic is called
\emph{sufficient} for a parameter in a given model if no other statistic
computed from the data, including the data themselves, provides any additional
information to the parameter's estimation.)

A natural way of representing experimental variables, auxiliary variables and
their relationships is through graphical models, and~in particular Bayesian
networks (BNs) \cite{koller}. BNs represent variables as nodes in a directed
acyclic graph in which arcs encode probabilistic dependencies. The~graphical
component of BNs allows for automated probabilistic manipulations via belief
propagation, which in turn makes it possible to compute the marginal and
conditional distributions of experimental variables in the presence of
incomplete data. However, learning a BN from data, that is, learning its
graphical structure and parameters, is a challenging problem; we refer the
reader to~\cite{neerlandica19} for a recent review on the~topic.

The EM algorithm can be used in its original form to learn the parameters of a
BN. The~{Structural EM} algorithm~\cite{sem1,sem2} builds on the EM algorithm to
implement structure learning: it computes the sufficient statistics required to
score candidate BNs in the E-step. However, the~Structural EM is computationally
demanding due to the large number of candidate models that are evaluated in the
search for the optimal BN. Using EM for parameter learning can be
computationally demanding as well for medium and large BNs. Hence, practical
implementations, e.g., \cite{bnstruct,cassio-kmax}, of both often replace belief
propagation with single imputation: each missing value is replaced with its
expected value conditional on the values observed for the other variables in the
same observation. This is a form of \emph{hard EM} because we are making hard
assignments of values; whereas using belief propagation would be a form of
\emph{soft EM}. This change has the effect of voiding the theoretical
guarantees for the Structural EM in~\cite{sem1,sem2}, such as the consistency of
the algorithm. Hard EM, being a form of single imputation, is also known to be
problematic for learning the parameters of statistical models~\cite{schafer}.

In this paper, we investigate the impact of learning the parameters and the
structure of a BN using hard EM instead of soft EM with a comprehensive
simulation study covering incomplete data with a wide array of different
characteristics. All the code used in the paper is available as an
R~\cite{rcore} package
(\url{https://github.com/madlabunimib/Expectation-Maximisation}).

The rest of the paper is organised as follows.  In~Section~\ref{sec:methods}, we
will introduce BNs (Section~\ref{sec:bns}); missing~data
(Section~\ref{sec:missing}); imputation (Section~\ref{sec:imputation}); and the
EM algorithm (Section~\ref{sec:em}). We describe the experimental setup we use
to evaluate different EM algorithms in the context of BN learning in
Section~\ref{sec:materials}, and~we report on the results in
Section~\ref{sec:results}. Finally, we provide practical recommendations on when
to use each EM algorithm in BN structure learning in
Section~\ref{sec:conclusions}. Additional details on the experimental setup are
provided in the~appendix.

\section{Methods}
\label{sec:methods}

This section introduces the notation and key definitions for BNs and incomplete
data. We then discuss possible approaches to learn BNs from incomplete data,
focusing on the EM and Structural EM~algorithms.

\subsection{Bayesian~Networks}
\label{sec:bns}

A Bayesian network BN \cite{koller} is a probabilistic graphical model that
consists of a directed acyclic graph (DAG) $\G = (\V, E)$ and a set of random
variables over $\X = \{X_1, \ldots, X_N\}$ with parameters $\T$. Each node in
$\V$ is associated with a random variable in $\X$, and~the two are usually
referred to interchangeably. The directed arcs $E$ in $\G$ encode the
conditional independence relationships between the random variables using
graphical separation, which is called \emph{d-separation} in this context.
As~a result, $\G$ leads to the decomposition \begin{equation} \Prob(\X \given
\G, \T) = \prod_{i=1}^{N} \Prob\left(X_{i} \given \Pai, \Ti \right),
\label{eq:parents} \end{equation} in which the global (joint) distribution of
$\X$ factorises in a local distribution for each $X_i$ that is conditional on
its parents $\Pai$ in $\G$ and has parameters $\Ti$. In~this paper, we assume
that all random variables are categorical: both $\X$ and the $X_i$ follow
multinomial distributions, and~the parameters $\Ti$ are conditional probability
tables (CPTs) containing the probability of each value of $X_i$ given each
configuration of the values of its parents $\Pai$. In~other words,
\begin{equation*} \Ti = \left\{ \piijk , k = 1, \ldots, |X_i|, j = 1, \ldots,
|\Pai| \right\} \end{equation*} where $\piijk = \Prob(X_i = k \given \Pai = j)$.
This class of BNs is called \emph{discrete BNs} \cite{heckerman}.
Other~classes commonly found in the literature include Gaussian BNs (GBNs)
\cite{heckerman3}, in~which $\X$ is a multivariate normal random variable and
the $X_i$ are univariate normals linked by linear dependencies; and~conditional
Gaussian BNs (CLGBNs) \cite{lauritzen}, which combine categorical and normal
random variables in a mixture~model.

The task of learning a BN from a data set $\D$ containing $n$ observations is
performed in two steps:
\begin{equation*}
  \underbrace{\Prob(\G, \T \given \D)}_{\text{learning}} =
    \underbrace{\Prob(\G \given \D)}_{\text{structure learning}} \cdot
    \underbrace{\Prob(\T \given \G, \D)}_{\text{parameter learning}}.
\end{equation*}

{Structure learning} consists of finding the DAG $\G$ that encodes the
dependence structure of the data, thus maximising $\Prob(\G \given \D)$ or some
alternative goodness-of-fit measure; \emph{parameter learning} consists in
estimating the parameters $\T$ given the $\G$ obtained from structure learning.
If~we assume that different $\Ti$ are independent and that data are complete
\citep{heckerman}, we can perform parameter learning independently for each node
following \mref{eq:parents} because
\begin{equation}
  \Prob(\T \given \G, \D) = \prod_{i=1}^N \Prob\left(\Ti \given \Pai, \D\right).
  \label{eq:decomp}
\end{equation}

Furthermore, if~$\G$ is sufficiently sparse, each node will have a small number
of parents; and~$X_i \given \Pai$ will have a low-dimensional parameter space,
making parameter learning computationally efficient. Both structure and
parameter learning involve $\Ti$, which can be estimated as using maximum
likelihood
\begin{equation}
  \hpiijk = \frac{n_{ijk}}{\sum_k n_{ijk}}
  \label{eq:thetafreq}
\end{equation}
or Bayesian posterior estimates
\begin{equation}
  \hpiijk = \frac{\aijk + n_{ijk}}{\sum_k \aijk + n_{ijk}}
  \label{eq:thetabayes}
\end{equation}
where the $\aijk > 0$ are hyperparameters of the (conjugate) Dirichlet prior for
$X_i \given \Pai$ and the $n_{ijk}$ are the corresponding counts computed from
the data. Estimating the $\Ti$ is the focus of parameter learning but they are
also estimated in structure learning. Directly, when $\Prob(\G \given \D)$ is
approximated with the Bayesian Information Criterion (BIC) \cite{schwarz}:
\begin{equation}
  \mathrm{BIC}(\G, \Theta \given \D) =
    \sum_{i = 1}^{N} \log \Prob\left(X_i \given \Pai, \Ti\right) -
      \frac{\log(n)}{2} |\Ti|.
  \label{eq:bic}
\end{equation}

Indirectly, through the $\{ \aijk + n_{ijk} \}$, we are computing
$\Prob(\G \given \D)$ as
\begin{align}
  &\Prob(\G \given \D) \propto \Prob(\G) \Prob(\D \given \G)&
  &\text{with}&
  &\Prob(\D \given \G) =
  \prod_{i=1}^N \prod_{j = 1}^{|\Pai|}
    \left[
      \frac{\Gamma(\sum_k \aijk)}{\Gamma(\sum_k \aijk + n_{ijk})}
      \prod_{k=1}^{|X_i|} \frac{\Gamma(\aijk + n_{ijk})}{\Gamma(\aijk)}
    \right].
  \label{eq:bd}
\end{align}

The expression on the right is the marginal probability of the data given a DAG
$\G$ averaging over all possible $\Ti$, and~is known as the Bayesian Dirichlet
score (BD) \cite{heckerman}. In~all of \mref{eq:thetafreq}--\mref{eq:bd} depends
on the data only through the counts $\{ n_{ijk} \}$, which are the
\emph{minimal sufficient statistics} for estimating all the quantities~above.

Once both $\G$ and $\T$ have been learned, we can use the BN to answer queries
about our quantities of interest using either \emph{exact} or
\emph{approximate inference} algorithms that work directly on the
BN~\cite{crc13}. Common choices are \emph{conditional probability} queries,
in~which we compute the posterior probability of one or more variables given the
values of other variables; and \emph{most probable explanation} queries,
in~which we identify the configuration of values of some variables that has the
highest posterior probability given the values of some other variables.
The~latter are especially suited to implement both prediction and imputation of
missing data, which we will discuss~below.

\subsection{Missing~Data}
\label{sec:missing}

A data set $\D$ comprising samples from the random variables in $\X$ is called
\emph{complete} when the values of all $X_i$ are known, that is, observed,
for~all samples. On~the other hand, if~$\D$ is \emph{incomplete}, some samples
will be completely observed while others will contain missing values for some
$X_i$. The~patterns with which data are missing are called the \emph{missing
data mechanism}. Modelling these mechanisms is crucial because the properties of
missing data depend on their nature, and~in particular, on~whether the fact that
values are missing is related to the underlying values of the variables in the
data set. The~literature groups missing data mechanisms in three
classes~\cite{rubin2}:
\begin{itemize}
  \item \emph{Missing completely at random} (MCAR): missingness does not
    depend on the values of the data, missing or observed.
  \item \emph{Missing at random} (MAR): missingness depends on the variables
    in $\X$ only through the observed values in the data.
  \item \emph{Missing not at random} (MNAR): the missingness depends on both
    the observed and the missing values in the data.
\end{itemize}

MAR is a sufficient condition for likelihood and Bayesian inference to be the
valid without modelling the missing data mechanism explicitly; hence, MCAR and
MAR are said to be \emph{ignorable patterns} of~missingness.

In the context of BNs, we can represent the missing data mechanism for each
$X_i$ with a binary latent variable $Z_i$ taking value $0$ for an observation if
$X_i$ is missing and $1$ otherwise. $Z_i$ is included in the BN as an additional
parent of $X_i$, so that the local distribution of $X_i$ in \mref{eq:parents}
can be written as
\begin{equation*}
  X_{i} \given \Pai, Z_i \sim \left\{
  \begin{array}{ll}
    \Prob^{(O)}\left(X_{i} \given \Pai, \Ti^{(O)}\right) & \text {for $Z_i = 1$} \\[1em]
    \Prob^{(M)}\left(X_{i} \given \Pai, \Ti^{(M)}\right) & \text {for $Z_i = 0$}
  \end{array}
  \right..
\end{equation*}

In the case of MCAR, $Z_i$ will be independent from all the variables in $\X$
except $X_i$; in the case of MAR, $Z_i$ can depend on $X_j \given Z_j = 1$ ($Z_i
\indep X_j \given Z_j = 1$) but not on $X_j \given Z_j = 0$ ($Z_i \notindep X_j
\given Z_j = 0$) for all $j \neq i$. In~both cases, the~data missingness is
ignorable, meaning that
\begin{equation*}
  \Prob^{(O)}\left(X_{i} \given \Pai, \Ti^{(O)}\right) =
  \Prob^{(M)}\left(X_{i} \given \Pai, \Ti^{(M)}\right) =
    \Prob\left(X_{i} \given \Pai, \Ti\right).
\end{equation*}

However, this is not the case for MNAR, where the missing data mechanism must be
modelled explicitly for the model to be learned correctly and for any inference
to be valid. As~a result, different~approaches to handle missing data will be
effective depending on which missing data mechanism we assume for the~data.

\subsection{Missing Data~Imputation}
\label{sec:imputation}

A possible approach to handle missing data is to transform an incomplete data
set into a complete one. The~easiest way to achieve this is to just remove all
the observations containing at least one missing value. However, this can
markedly reduce the amount of data and it is widely documented to introduce bias
in both learning and inference, see, for~instance \cite{jahdav,raghunathan}.
A~more principled approach is to perform \emph{imputation}; that is,
to~predict missing values based on the observed ones. Two groups of imputation
approaches have been explored in the literature: \emph{single imputation} and
\emph{multiple imputation}. We~provide a quick overview of both below, and~we
refer the reader to~\cite{rubin2} for a more comprehensive
theoretical~treatment.

\emph{Single imputation} approaches impute one value for each missing value in
the data set. \mbox{As a result,} they can potentially introduce bias in
subsequent inference because the imputed values naturally have a smaller
variability than the observed values. Furthermore, it is impossible to assess
imputation uncertainty from that single value. Two examples of this type of
approach are mean imputation (replacing each missing value with the sample mean
or mode) or k-NN imputation~\cite{knn,knn2} (replacing~each missing value with
the most common value from similar cases identified via $k$-nearest~neighbours).

\emph{Multiple imputation} replaces each missing value by $B$ possible values
to create $B$ complete data sets, usually with $B \in [5, 10]$. Standard
complete-data probabilistic methods are then used to analyse each completed data
set, and~the $B$ completed-data inferences are combined to form a single
inference that properly reflects uncertainty due to missingness under that
model~\cite{schafer}. A~popular example is multiple imputation by chained
equation~\cite{chained}, which has seen widespread use in the medical and
life~sciences fields.

It is important to note that there are limits to the amount of missing data that
can be effectively managed. While there is no common guideline on that, since
each method and missing data mechanism are different in that respect, suggested
limits found in the literature range from 5\% \cite{schafer} to 10\%
\cite{bennett}.

\subsection{The Expectation-Maximisation (EM) Algorithm}
\label{sec:em}

The imputation methods described above focus on completing individual missing
values with predictions from the observed data without considering what
probabilistic models will be estimated from the completed data. The
Expectation-Maximisation (EM) \cite{em} algorithm takes the opposite view:
starting from the model we would like to estimate, it identifies the sufficient
statistics we need to estimate its parameters and it completes those sufficient
statistics by averaging over the missing values. The~general nature of this
formulation makes EM applicable to a wide range of probabilistic models,
as~discussed in~\cite{watanabe,mclachlan} as well as~\cite{rubin2}.

EM (Algorithm \ref{alg:em-soft}) is an iterative algorithm consisting of the
following two~steps:
\begin{itemize}
  \item the \emph{Expectation} step (E-step) consists in computing the
    expected values of the sufficient statistics $s(\D)$ for the parameters
    $\T_j$ using the previous parameter estimates $\hT_{j-1}$;
  \item the \emph{Maximisation} step (M-step) takes the sufficient statistics
    $\widehat{s}_j$ from the E-step and uses them to update the parameters
    estimates.
\end{itemize}

Both maximum likelihood and Bayesian posterior parameter estimates are allowed
in the M-Step. The~E-step and the M-step are repeated until convergence. Each
iteration increases marginal likelihood function for the observed data, so the
EM algorithm is guaranteed to converge because the marginal likelihood of the
(unobservable) complete data is finite and bounds above that of the
current~model.

\pagebreak

\begin{algorithm}[H]
\SetAlgoLined
\DontPrintSemicolon
  Choose an initial value $\hT_0$ for $\T$.\;
  \While{$\left|\hT_{j-1} - \hT_{j}\right| < \varepsilon$,
         iterating over $j = 1, 2, \ldots$ :}{
    \textbf{Expectation step}: compute the expected sufficient statistics for
      $\T$ over both the observed and   missing data, conditional on the current
      estimate of $\T$:
      \begin{equation*}
        \widehat{s}_{j} = \E\left(s(\D) \given \hT_{j-1}\right)
      \end{equation*}
    \textbf{Maximisation step}: compute the new estimate $\hT_{j}$ from
      $\widehat{s}_{j}$.
  }
  Estimate $\T$ with the last $\hT_{j}$.
  \caption{The (Soft) Expectation-Maximisation Algorithm (Soft EM)}
  \label{alg:em-soft}
\end{algorithm}
\vspace{\baselineskip}

One key limitation of EM as described in Algorithm \ref{alg:em-soft} is that the
estimation of the expected sufficient statistics may be computationally
unfeasible or very costly, thus making EM impractical for use in real-world
applications. As~an alternative, we can use what is called {hard EM}, which is
shown in Algorithm \ref{alg:em-hard}. Unlike the standard EM, hard EM computes
the expected sufficient statistics $s(\D)$ by imputing the missing data
$\D^{(M)}$ with their most likely completion $c(\cdot)$, and~then using the
completed data $\widehat{\D}$ to compute the sufficient statistics. Hence the
name, we replace the missing data with hard assignments. In~contrast,
the~standard EM in Algorithm \ref{alg:em-soft} is sometimes called {soft EM}
because it averages over the missing values, that is, it considers all its
possible values weighted by their probability~distribution.
\vspace{\baselineskip}

\begin{algorithm}[H]
\SetAlgoLined
\DontPrintSemicolon
  Choose an initial value $\hT_0$ for $\T$.\;
  \While{$\left|\hT_{j-1} - \hT_{j}\right| < \varepsilon$,
         iterating over $j = 1, 2, \ldots$:}{
    \textbf{Expectation step}: impute the missing data with their expectations
      to create the completed data set
      \begin{equation*}
        \widehat{\D}_j = \left\{ \D^{(O)},\; \widehat{D}^{(M)}
                       = c\left(\D^{(M)} \given \hT_{j-1}\right) \right\}
      \end{equation*}
      and then compute the sufficient statistics for $\T$ as
      \begin{equation*}
        \widehat{s}_j = s\left(\D_j\right)
      \end{equation*}
    \textbf{Maximisation step}: compute the new estimate $\hT_{j}$ from
      $\widehat{s}_{j}$.
  }
  Estimate $\T$ with the last $\hT_{j}$.
  \caption{The Hard EM~Algorithm.}
  \label{alg:em-hard}
\end{algorithm}
\vspace{\baselineskip}

It is important to note that hard EM and soft EM, while being both formally
correct, may display very different behaviour and convergence rates. Both
algorithms behave similarly when the random variables that are not completely
observed have a skewed distribution~\cite{koller}.

\subsection{The EM Algorithm and Bayesian~Networks}
\label{sec:embn}

In the context of BNs, EM can be applied to both parameter learning and
structure learning. For~the parameter learning, the~E-step and M-step~become:
\begin{itemize}
  \item the \emph{Expectation} (E) step consists of computing the expected
    values of the sufficient statistics (the~counts $\{ n_{ijk} \}$) using exact
    inference along the lines described above to make use of incomplete as well
    as complete samples;
  \item the \emph{Maximisation} (M) step takes the sufficient statistics from
    the E-step and estimates the parameters of the BN.
\end{itemize}
As for structure learning, the~Structural EM algorithm \citep{sem1} implements
EM as~follows:
\begin{itemize}
  \item in the E-step, we complete the data by computing the expected sufficient
    statistics using the current network structure;
  \item in the M-step, we find the structure that maximises the expected score
    function for the completed~data.
\end{itemize}

This approach is computationally feasible because it searches for the best
structure inside of the EM, instead of embedding EM inside a structure learning
algorithm; and it maintains the convergence guarantees of the original both in
its maximum likelihood~\cite{sem1} and Bayesian~\cite{sem2} formulations.
However, the~Structural EM is still quite expensive because of the large number
and the dimensionality of the sufficient statistics that are computed in each
iteration.~\cite{sem1} notes: ``Most of the running time during the execution of
our procedure is spent in the computations of expected statistics. This is where
our procedure differs from parametric EM. In~parametric EM, we know in advance
which expected statistics are required. [\ldots] In our procedure, we cannot
determine in advance which statistics will be required. Thus, we have to handle
each query separately. Moreover, when we consider different structures, we~are
bound to evaluate a larger number of queries than parametric EM.'' Even if we
explore candidate DAGs using a local search algorithm such as hill-climbing,
and~even if we only score DAGs that differ from the current candidate by a
single arc, this means computing $O(N)$ sets of sufficient statistics.
Furthermore, the~size of each of these sufficient statistics increases
combinatorially with the size of the $\Pai$ of the corresponding $X_i$.

As a result, practical software implementations of the Structural EM such as
those found in~\cite{bnstruct,cassio-kmax} often replace the soft EM approach
described in Algorithm \ref{alg:em-soft} with the hard EM from
Algorithm~\ref{alg:em-hard}. The~same can be true for applications of EM to
parameter learning, for~similar reasons: the cost of using exact inference can
become prohibitive if $\X$ is large or if there is a large number of missing
values. Furthermore, the~decomposition in \mref{eq:decomp} no longer holds
because the expected sufficient statistics depend on all $X_i$. In practice this
means that, instead of computing the expected values of the $\{ n_{ijk} \}$ as
weighted average over all possible imputations of the missing values, we perform
a single imputation of each missing value and use the completed observation to
tally up $\{ n_{ijk} \}$.

This leads us to the key question we address in this paper: what is the impact
of replacing soft EM with hard EM on learning BNs?

\section{Materials}
\label{sec:materials}

In order to address the question above, we perform a simulation study to compare
soft and hard EMs. We limit ourselves to discrete BNs, for~which we explore both
parameter learning (using a fixed gold-standard network structure) and structure
learning (using network structures with high $\Prob(\G \given \D)$ that would be
likely candidate BNs during learning). In~addition, we consider a variant of
soft EM in which we use early stopping to match its running time with that of
hard EM. We will call it \emph{soft-forced EM}, meaning that we force it to
stop without waiting for it to converge. In~particular, \emph{soft-forced EM}
stops after 3, 4 and 6 iterations, respectively, for \emph{small},
\emph{medium} and \emph{large} networks.

The study investigates the following experimental~factors:
\begin{itemize}
  \item Network size: small (from 2 to 20 nodes), medium (from 21 to 50 nodes)
    and large (more than 50~nodes).
  \item Missingness balancing: whether the distribution of the missing values
    over the possible values taken by a node is balanced or unbalanced (that
    is, some values are missing more often than others).
  \item Missingness severity: low ($\leqslant$1\% missing values), medium (1\%
    to 5\% missing values) and high (\mbox{5\% to 20\% missing values}).
  \item Missingness pattern: whether missing values appear only in root nodes
    (labelled ``root''), only in leaf nodes (``leaf''), in~nodes with large
    number of neighbours (``high degree'') or uniformly on all node types
    (``fair''). We also consider specific target nodes that represent the
    variables of interest in the BN (``target'').
  \item Missing data mechanism: the \emph{ampute} function of the
    \textbf{mice} R package~\cite{amputepackage} has been applied to generated
    data sets to simulate MCAR, MAR and~MNAR missing data mechanisms as
    described in Section~\ref{sec:missing}.
\end{itemize}

We recognise that these are but a small selection of the characteristics of the
data and of the missingness patterns that might determine differences in the
behaviour of soft EM and hard EM. We~focus on these particular experimental
factors because either they can be empirically verified from data, or~they must
be assumed in order to learn any probabilistic model at all, and~therefore
provide a good foundation for making practical recommendations for real-world
data~analyses.

The simulation study is based on seven reference BNs from \emph{The Bayesys
data and Bayesian network repository} \cite{repository}, which are summarised in
Table~\ref{tab:considerdataset}. We generate incomplete data from each of them
as~follows:
\begin{enumerate}[leftmargin=*,labelsep=5.8mm]
  \item We generate a complete data set from the BN. \label{sim:step1}
  \item We introduce missing values in the data from step \ref{sim:step1} by
    hiding a random selection of observed values in a pattern that satisfies the
    relevant experimental factors (missingness balancing, missingness severity,
    missingness pattern and missing data mechanism). We perform this step 10
    times for each complete data set. \label{sim:step2}
  \item We check that the proportion of missing values in each incomplete data
    set from step \ref{sim:step2} is within a factor of 0.01 of the missingness
    severity. \label{sim:step3}
  \item We perform parameter learning with each EM algorithm and each incomplete
    data set to estimate the $\hT_{i}$ for each node $X_i$, which we then use to
    impute the missing values in those same data sets. As~for the network
    structure, we consider both the DAG of the reference BN and a set network
    structures with high $\Prob(\G \given \D)$. \label{sim:step4}
\end{enumerate}

\vspace{-0.5\baselineskip}
\begin{table}[H]
\caption{Reference Bayesian networks (BNs) used to generate the data in the simulation~study.}
\centering
\begin{tabular}{ccc}
  \toprule
  \textbf{Network's Size} & \textbf{Bayesian Network} & \textbf{Number of Nodes} \\
  \midrule
  \multirow{2}{*}{\text{\begin{tabular}[c]{@{}c@{}}small\\ (from 2 to 20 nodes)\end{tabular}}} & \emph{Asia} & 8 \\
  \cline{2-3}
  & \emph{Sports} & 9  \\
  \midrule
  \multirow{2}{*}{\text{\begin{tabular}[c]{@{}c@{}}medium \\ (from 21 to 50 nodes)\end{tabular}}} & \emph{Alarm} & 31  \\
  \cline{2-3}
  & \emph{Property} & 27 \\
  \midrule
  \multirow{3}{*}{\text{\begin{tabular}[c]{@{}c@{}}large \\ (more than 50 nodes)\end{tabular}}} & \emph{Hailfinder} & 56 \\
  \cline{2-3}
  & \emph{Formed} & 88  \\
  \cmidrule{2-3}
  & \emph{Pathfinder} & 109  \\
  \bottomrule
\end{tabular}%
\label{tab:considerdataset}
\end{table}

The complete list of simulation scenarios is included Appendix \ref{appendixA}.

We measure the performance of the EM algorithms~with:
\begin{itemize}
  \item The \emph{proportion of correct replacements} (PCR), defined as the
    number of missing values that are correctly replaced. Higher values are
    better.
  \item The \emph{absolute probability difference}:
    \begin{equation}
      \operatorname{APD} = \sum_{m = 1}^{M} |p_m - q_m|,
    \end{equation}
    where $M$ is the number of missing values; $p_m$ is the probability of the
    $m$th missing value computed using the reference BN; and $q_m$ is the
    probability of the $m$th missing value computed using the EM algorithm.
    Lower values are better.
  \item The \emph{Kullback--Leibler divergence}:
    \begin{align}
      &\KL\left[\Theta \klsep \hT\right]
          = \sum_{m = 1}^{M} \KL\left[\Theta_{(m)} \klsep \hT_{(m)} \right]&
      &\text{where}&
      &\KL\left[\Theta_{(m)} \klsep \hT_{(m)} \right] =
          \sum \Theta_{(m)} \log \frac{\Theta_{(m)}}{\hT_{(m)}},
    \end{align}
    where the $\Theta_{(m)}$ are the conditional probabilities for the random
    variable of the $m$th missing value computed using reference BN; and the
    $\hT_{(m)}$ are the corresponding conditional probabilities computed by the
    EM algorithm. Lower values are better.
\end{itemize}

These three measures compare the performance of different EM algorithms at
different levels of detail. PCR provides a rough indication about the overall
performance in terms of how often the EM algorithm correctly imputes a missing
value, but~it does not give any insight on how incorrect imputations occur.
Hence, we also consider APD and KLD, which measure how different are the
$\hT_{i}$ produced by each EM algorithm from the corresponding $\Ti$ in the
reference BN. These two measures have a very similar behaviour in our simulation
study, so for brevity we will only discuss~KLD.

As mentioned in step \ref{sim:step4}, we compute the performance measures using
both the network structure of the reference BN and a set of network structures
with high $\Prob(\G \given \D)$. When using the former, which~can be taken as an
``optimal'' structure, we are focusing on the performance of EM  as it would be
used in parameter learning. When using the latter, we are instead focusing on EM
as it would be used in the context of a structure learning algorithm like the
Structural EM. Such an algorithm will necessarily visit other network structures
with high $\Prob(\G \given \D)$ while looking for an optimal one. Such~networks
will be similar to that of the reference BN; hence, we generate them by
perturbing its structure by removing, adding or reversing a number of arcs.
In~particular:
\begin{enumerate}[leftmargin=*,labelsep=5.8mm]
  \item we choose to perturb 15\% of nodes in small BNs and 10\% of nodes in
    medium and large BNs, to~ensure a fair amount of perturbation across BNs of
    different size;
  \item we sample the nodes to perturb;
  \item and then we apply, to~each node, a~perturbation chosen at random among
    \emph{single~arc removal}, \emph{single~arc addition} and \emph{single
    arc reversal}.
\end{enumerate}

We evaluate the performance of the EM algorithms with the perturbed networks
using
\begin{equation*}
  \Delta \KL = \KL\left[\Theta_\text{perturbed} \klsep \hT\right] -
               \KL\left[\Theta_\text{reference} \klsep \hT\right],
\end{equation*}
that is, the~difference in KLD divergence between the BN learned by EM from the
perturbed networks and reference BN, and~the KLD divergence between the BN
learned by EM from the network structure of the reference BN and the reference
BN itself. The~difference is evaluated on 10 times for each combination of
experimental factors on 10 different incomplete data sets. Lower values are
better because they suggest a good level agreement between
$\Theta_\text{perturbed}$ and $\Theta_\text{reference}$ and a small level of
information~loss.

\section{Results}
\label{sec:results}

\begin{figure}[H]
\begin{center}
  \includegraphics[width=\linewidth]{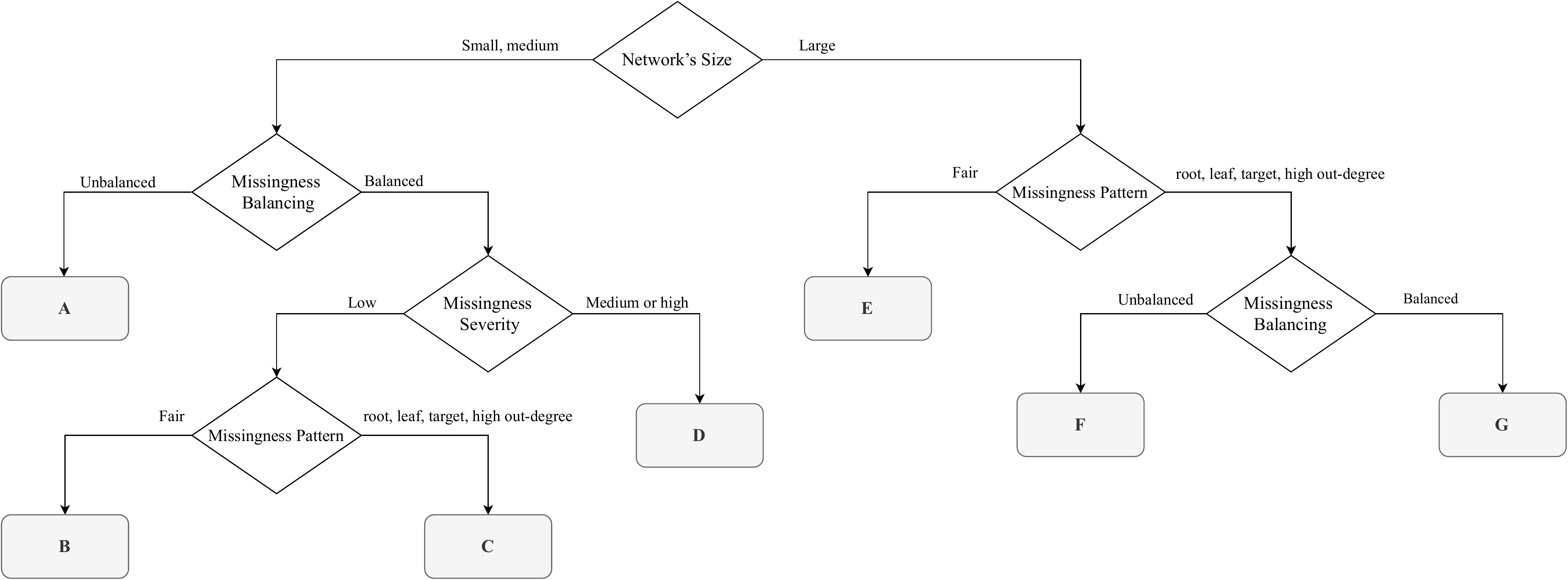}
  \caption{Decision tree for best practice~guidance.}
  \label{fig:select-algorithm}
\end{center}
\end{figure}

The results, comprising 5520 incomplete data sets and the resulting BNs, are
summarised in Figure~\ref{fig:select-algorithm} for the simulations in which we
are using the network structure of the reference BNs. The~decision tree shown in
the Figure is intended to provide guidance to practitioners on which imputation
algorithm appears to provide the best performance depending on the
characteristics of their incomplete data problem. Each leaf in the decision tree
corresponds to a subset of the scenarios we examined, grouped~by the values of
the experimental factors, and~to a recommendation which EM algorithm has the
best average KLD values. (Recommendations are also shown in
Table~\ref{tab::casetable} for convenience, along~with the leaf label
corresponding the reference BNs). For brevity, we will discuss in detail on
leaves A, B, E and G, which result into different recommended algorithms (Table
\ref{tab::casetable}).

\begin{table}[H]
\centering
\caption{Recommended algorithm by decision tree~leaf.}
\label{tab::casetable}
\begin{tabular}{ccc}
  \toprule
  \textbf{Leaf}  & \textbf{Recommended Algorithm} & \textbf{Bayesian Network} \\
  \midrule
  A & Hard, Soft, Soft-Forced & \begin{tabular}[c]{@{}c@{}}ASIA \\ ALARM \end{tabular} \\
  \midrule
  B & Hard                    & \begin{tabular}[c]{@{}c@{}}SPORTS\\ PROPERTY \end{tabular} \\
  \midrule
  C & Soft, Soft-Forced       & \begin{tabular}[c]{@{}c@{}}SPORTS\\ PROPERTY \end{tabular} \\
  \midrule
  D & Hard                    & \begin{tabular}[c]{@{}c@{}}SPORTS\\ PROPERTY \end{tabular} \\
  \midrule
  E & Hard                    & \begin{tabular}[c]{@{}c@{}}FORMED\\ PATHFINDER\\ HAILFINDER \end{tabular} \\
  \midrule
  F & Hard                    & \begin{tabular}[c]{@{}c@{}}FORMED\\ PATHFINDER\\ HAILFINDER \end{tabular} \\
  \midrule
  G & Soft, Soft-Forced       & \begin{tabular}[c]{@{}c@{}}FORMED\\ PATHFINDER \end{tabular} \\
  \toprule
\end{tabular}
\end{table}

\begin{figure}[H]
\begin{center}
  \includegraphics[width=0.94\linewidth]{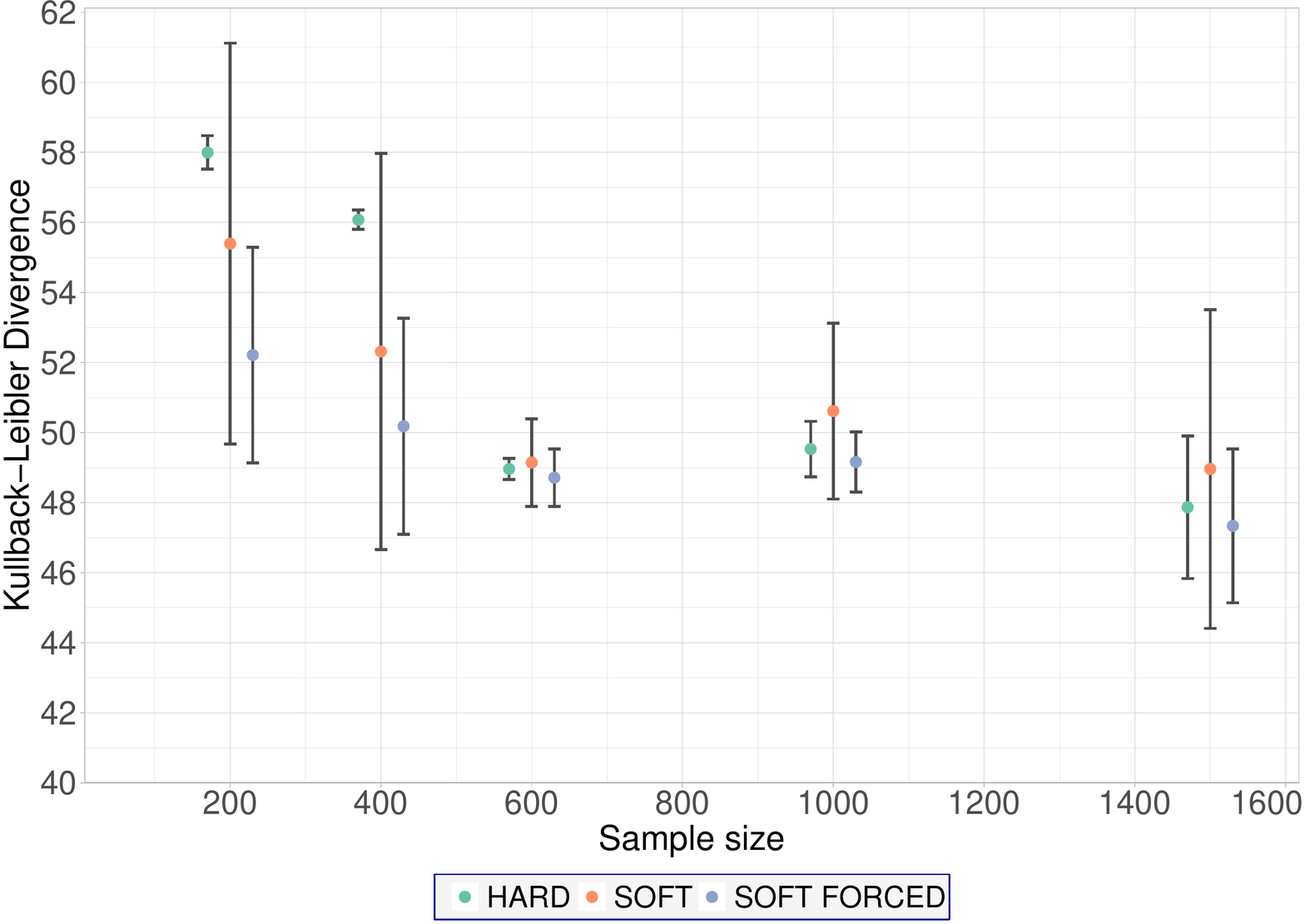}
  \caption{Leaf A. No EM algorithm proves to be more effective than the others
    (data sets with 5\% missing data generated from the \emph{Alarm} BN).}
  \label{fig:alarm-example}
\end{center}
\end{figure}

\begin{figure}[H]
\begin{center}
  \includegraphics[width=0.94\linewidth]{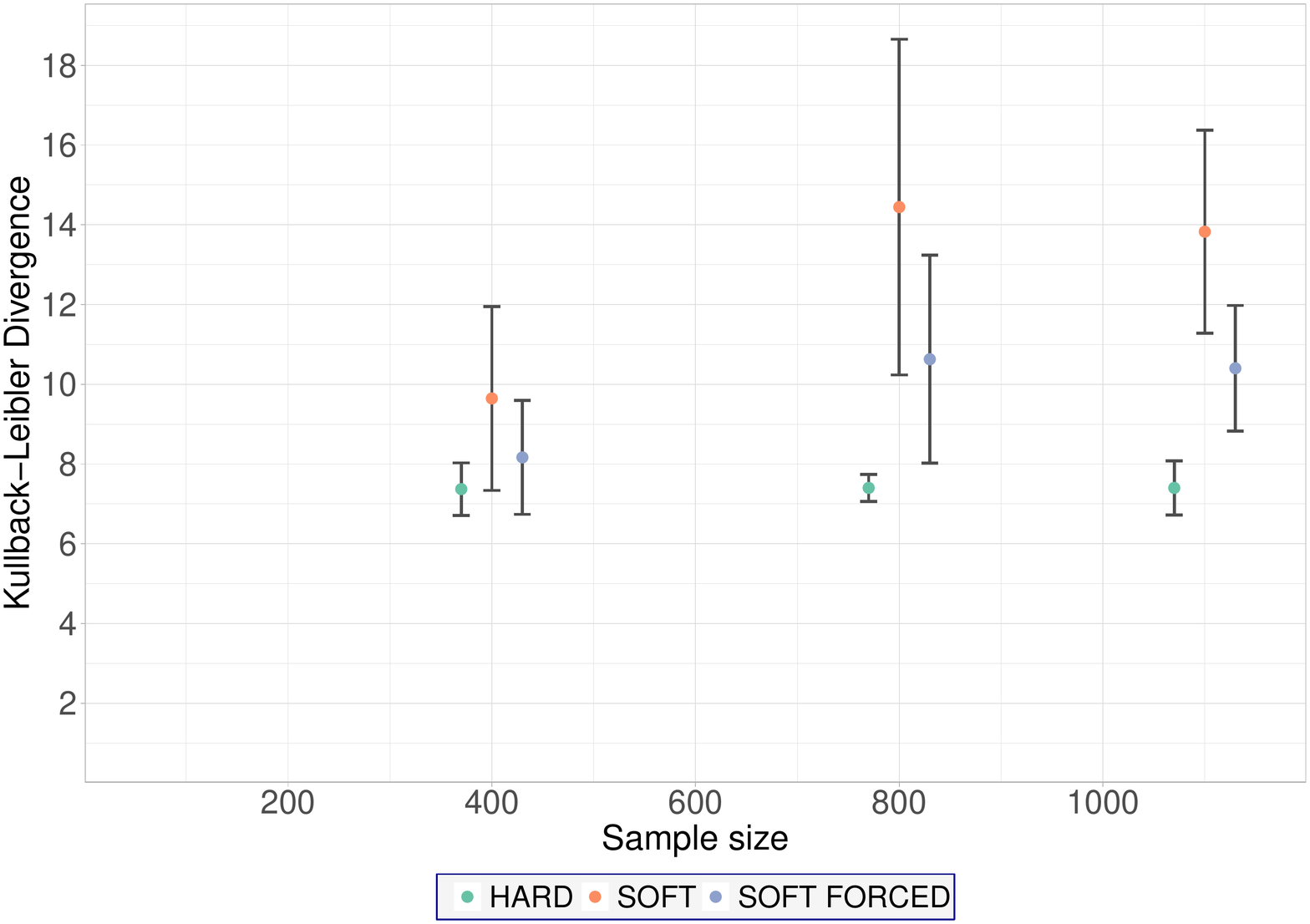}
  \caption{Leaf B. Hard EM achieves a value of KLD which is significantly
    smaller than that achieved by other EM algorithms (data sets with 5\%
    missing data generated from the \emph{Property} BN).}
  \label{fig:property-example}
\end{center}
\end{figure}

\begin{figure}[H]
\begin{center}
  \includegraphics[width=0.94\linewidth]{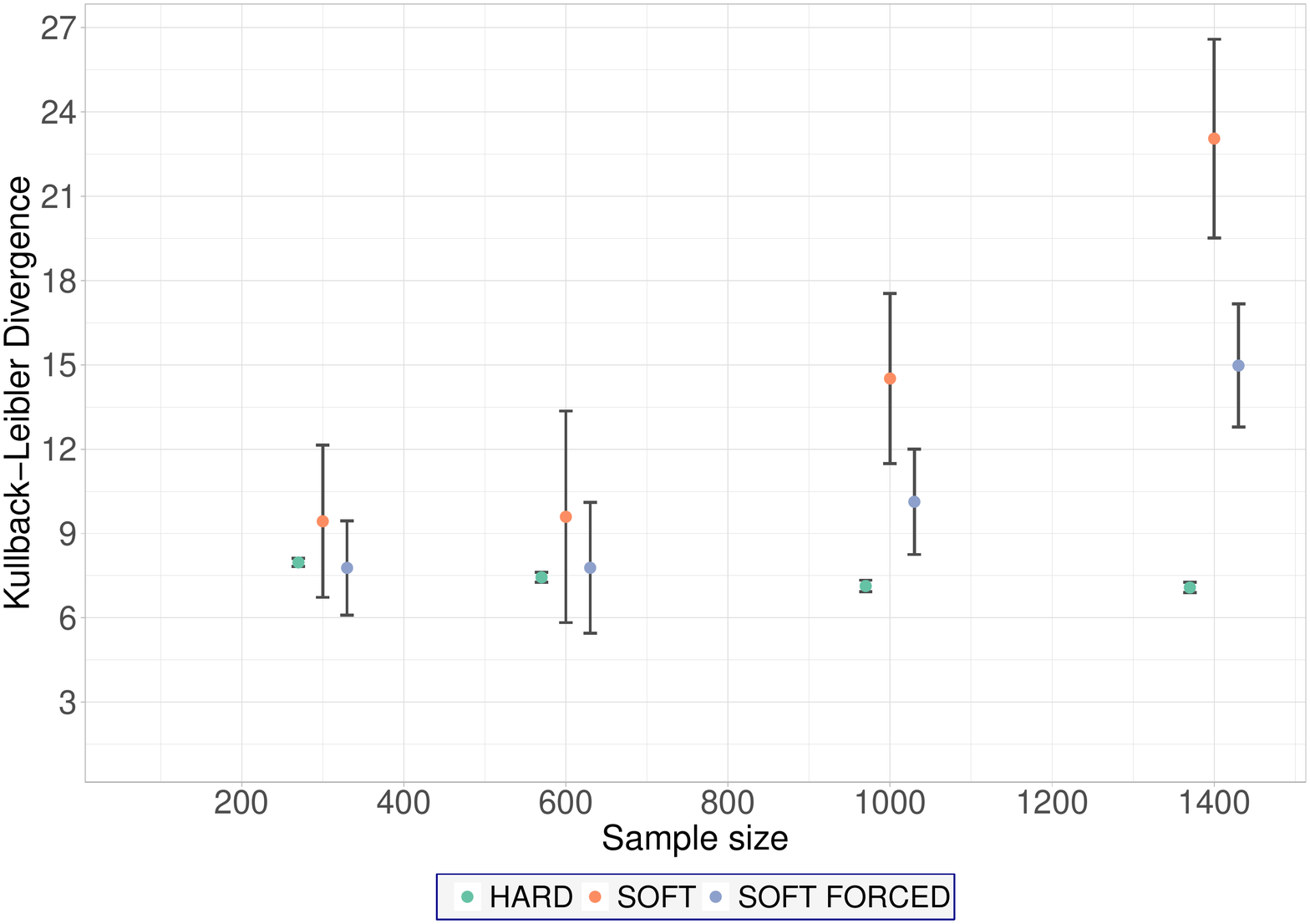}
  \caption{Leaf E. Hard EM achieves a value of KLD which is significantly
    smaller than that achieved by other EM algorithms (data sets with 1\%
    missing data generated from the \emph{Formed} BN).}
  \label{fig:formed-example}
\end{center}
\end{figure}

\begin{figure}[H]
\begin{center}
  \includegraphics[width=0.94\linewidth]{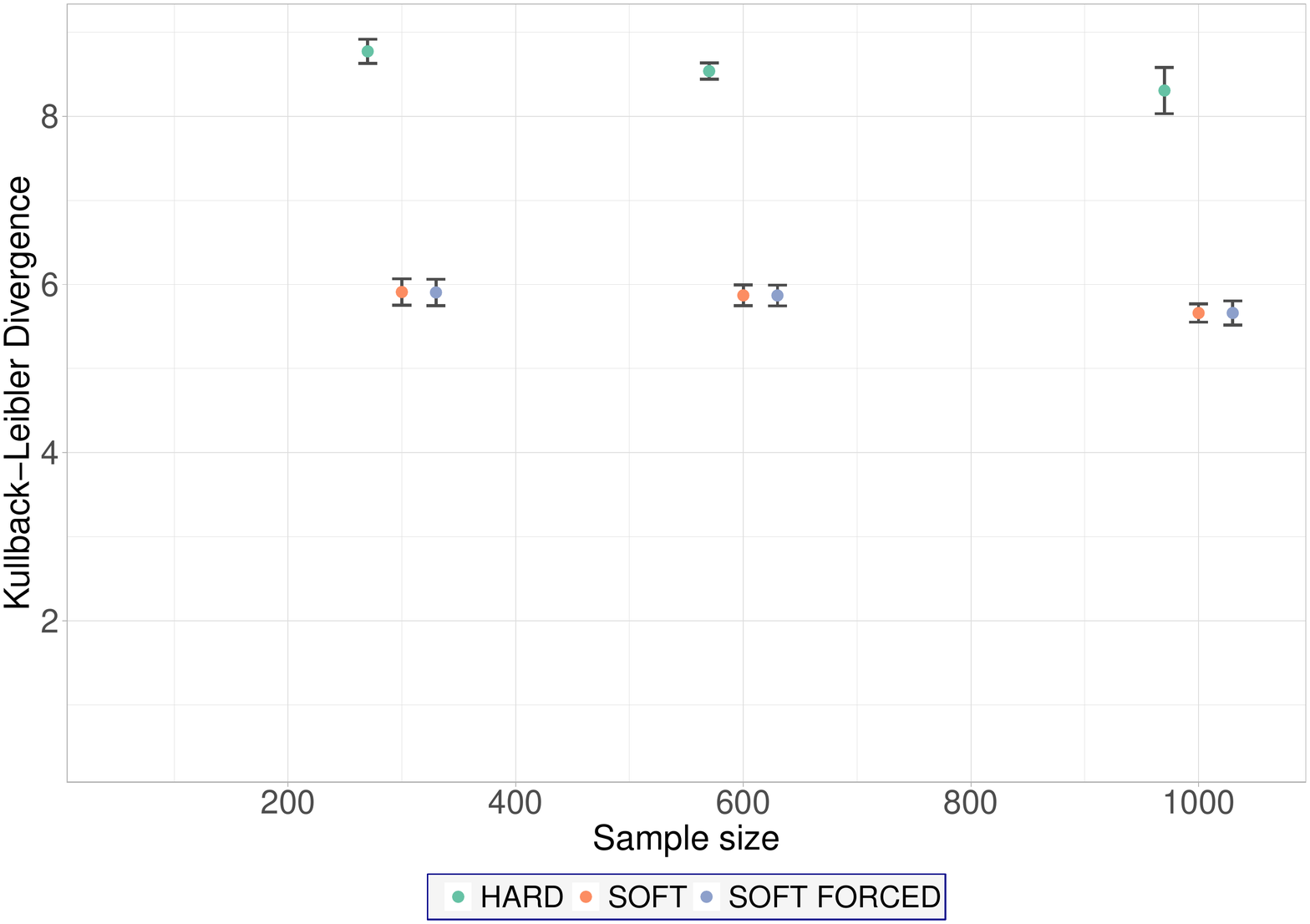}
  \caption{Leaf G. Hard EM achieves a value of KLD which is significantly
    greater than that achieved by other EM algorithms (data sets with 1\%
    missing data generated from the \emph{Pathfinder} BN).}
  \label{fig:pathfinder-example}
\end{center}
\end{figure}

The 95\% confidence intervals for KLD are shown in
Figures~\ref{fig:alarm-example}--\ref{fig:pathfinder-example}, respectively. We
use those confidence intervals to rank the performance of various EM algorithms:
we say an algorithm is better than another if it has a lower average KLD and
their confidence intervals do not~overlap.

Leaf A covers small and medium BNs in which variables have unbalanced
missingness distributions. In~these cases no EM algorithm dominates the others,
hence no specific algorithm is recommended. As~expected, KLD decreases as the
sample size increases for all~algorithms.

Leaf B covers small and medium BNs as well, but~in this case random variables
have balanced missingness distributions, the~frequency of missing values is low
and~the pattern of missingness is fair. In~these cases, hard EM is the
recommended algorithm (Figure \ref{fig:property-example}). It is important to
note that hard EM consistently has the lowest KLD value and has low variance,
while soft EM and soft-forced EM have a much greater variance even for large
samples~sizes.

Leaf E covers large BNs where the pattern of missingness is fair. In~these
cases, hard EM is the recommended algorithm (Figure \ref{fig:formed-example}).
Note that the performance of both the soft EM and the soft-forced EM algorithms
markedly degrades as the sample size increases. However, the~performance of hard
EM remains constant as the sample size~increases.

Leaf G covers large BNs where the pattern of missingness is not fair (the
missingness pattern is one of ``root'', ``leaf'', ``target'', ``high
degree''), and~the random variables have balanced missingness distributions.
In~these cases, soft EM is the recommended algorithm (Figure
\ref{fig:pathfinder-example}). The performance of all EM algorithms shows only
marginal improvements as the sample size increases, but~low~variance.

As for the remaining leaves, we recommend the hard EM algorithm for leafs D and
F. Leaf D covers small and medium BNs with balanced missingness distributions
and medium or high missingness severity; leaf F covers only large BNs and
unbalanced missingness. Finally, leaf C recommends soft and soft-forced EM for
small and medium BNs with balanced missingness distributions, low missingness
severity and a pattern of missingness that is not~fair.

As for the simulations that are based on the perturbed networks, the~increased
heterogeneity of the results makes it difficult to give recommendations as
detailed as those above. We note, however, some overall~trends:
\begin{itemize}
  \item Hard EM has the lowest $\Delta \KL$ in 44/67 scenarios, compared to
    16/67 (soft EM) and 7/67 (soft-forced EM). Soft EM has the highest $\Delta
    \KL$ in 30/67 triplets, compared to 24/67 (soft-forced) and 13/67 (hard EM).
    Hence, hard EM can often outperform soft EM in the quality of estimated
    $\hT_i$, and~it appears to be the worst-performing only in a minority of
    simulations. The~opposite seems to be true for soft EM, possibly because it
    converges very slowly or it fails to converge completely in some
    simulations. The~performance of soft-forced EM appears to be not as good as
    that of hard EM, but~not as often the worst as that of soft EM.
  \item We observe some negative $\Delta \KL$ values for all EM algorithms: 7/67
    (hard EM), 8/67 (soft EM), 5/67 (soft-forced). They highlight how all EM
    algorithms can sometimes fail to converge and produce good parameter
    estimates for the network structure of the reference BN, but~not for the
    perturbed network structures.
  \item Hard EM has the lowest $\Delta \KL$ 13/30 times in small networks, 9/14
    in medium networks and 21/23 in large networks in a monotonically increasing
    trend. At~the same time, hard EM has the highest $\Delta \KL$ in 8/30 times
    in small networks, 4/14 in medium networks and 0/23 in large networks, in~a
    monotonically decreasing trend. This suggests that the performance of hard
    EM improves as the BNs increase in size: it provides the best $\hT_i$ more
    and more frequently, and~it is never the worst performer in large networks.
  \item Soft EM has the lowest $\Delta \KL$ in 12/30 times in small networks,
    5/14 in medium networks and 0/23 in large networks in a monotonically
    increasing trend. At~the same time, soft EM has the highest $\Delta \KL$ in
    7/30 times in small networks, 6/14 in medium networks and 17/23 in large
    networks, in~a monotonically increasing trend. Hence, we observe that soft
    EM is increasingly unlikely to be the worst performer as the size of the BN
    increases, but~it is also increasingly likely outperformed by hard EM.
  \item Soft-forced EM never has the lowest $\Delta \KL$ in medium and large
    networks. It has the highest $\Delta \KL$ 15/30 times in small networks,
    4/14 in medium networks and 4/23 in large networks, in~a monotonically
    decreasing trend (with a large step between small and medium networks,
    and~comparable values for medium and large networks). Again, this suggests
    that the behaviour of soft-forced EM is an average of that of hard EM and
    soft EM, occupying the middle ground for medium and large networks.
\end{itemize}

\unskip

\section{Discussion and Conclusions}
\label{sec:conclusions}

BN parameter learning from incomplete data is typically performed using the EM
algorithm. Likewise, structure learning with the Structural EM algorithm embeds
the search for the optimal network structure within EM. Practical applications
of BN learning often choose to implement learning using the {hard EM} approach
(which is based on single imputation) instead of the {soft EM} approach (which
is based on belief propagation) for computational reasons despite the known
limitations of the former. To the best of the authors' knowledge, no previous
work has systematically compared hard EM to soft EM when applied to BN learning,
despite their popularity in several application fields. Hence, we investigated
the following question: what is the impact of using hard EM instead of soft EM
on the quality of the BNs learned from incomplete data? In addition, we also
considered an early-stopping variant of soft EM, which we called
\emph{soft-forced EM}. However, we find that it does not outperform hard EM or
soft EM in any simulation~scenario.

Based on a comprehensive simulation study, we find that in the case of
parameter~learning:
\begin{itemize}
  \item Hard EM performs well across BNs of different sizes when the missing
    pattern is fair; that is, missing data occur independently on the structure
    of the BN.
  \item Soft EM should be preferred to hard EM, across BNs of different sizes,
    when the missing pattern is  not fair; that is, missing data occur at nodes
    of the BN with specific graphical characteristics (root, leaf, high-degree
    nodes); and when the missingness distribution of nodes is balanced.
  \item Hard and soft EM perform similarly for medium-size BNs when missing data
    are unbalanced.
\end{itemize}

In the case of structure learning, which we explore by investigating a set of
candidate networks with high posterior probability, we find~that:
\begin{itemize}
  \item Hard EM achieves the lowest value of $\Delta \KL$ in most simulation
    scenarios, reliably outperforming other EM algorithms.
  \item In terms of robustness, we find no marked difference between soft EM and
    hard EM for small to medium BNs. On~the other hand, hard EM consistently
    outperforms soft EM for large BNs. In~fact, for~large BNs hard EM achieves
    the lowest value of $\Delta \KL$ in all simulations, and~it never achieves
    the highest value of $\Delta \KL$.
  \item Sometimes all EM algorithms fail to converge and to provide good
    parameter estimates for the network structure of the true BN, but~not for
    the corresponding perturbed networks.
\end{itemize}

However, it is important to note that this study presents two limitations.
Firstly, a~wider variety of numerical experiments should be performed to further
validate conclusions. The~complexity of capturing the key characteristics of
both BNs and missing data mechanisms make it extremely difficult to provide
comprehensive answers while limiting ourselves to a feasible set of experimental
factors. Secondly, we limited the scope of this paper to discrete BN: but
Gaussian BNs have seen wide applications in life sciences applications, and~it
would worthwhile to investigate to what extent our conclusions apply to them.
Nevertheless, we believe the recommendations we have collected in
Section~\ref{sec:results} and discussed here can be of use to practitioners
using BNs with incomplete~data.

\vspace{0.5\baselineskip}

\authorcontributions{Investigation, Andrea Ruggieri, Francesco Stranieri and
Marco Scutari; Methodology, Marco Scutari; Supervision, Fabio Stella; Writing –
original draft, Andrea Ruggieri and Francesco Stranieri; Writing – review \&
editing, Fabio Stella and Marco Scutari. }

\funding{This research received no external funding. }

\conflictsofinterest{The authors declare no conflict of~interest.}

\pagebreak

\appendixtitles{yes}

\appendix
\section{Complete List of the Simulation~Scenarios}
\label{appendixA}

In this appendix we provide a comprehensive list of all the experiments in the simulation study described in Section~\ref{sec:materials}, organised by their key characteristics in Table~\ref{tab:list-experiments}.

\begin{table}[H]

\caption{Complete description of all the combinations of experimental factors covered in the simulation~study.}

\begin{tabular}{ccccc}
\toprule
 & & \textbf{Proportion of} & & \\
\multirow{-2}{*}{\textbf{Network}} & \multirow{-2}{*}{\textbf{Description}} & \textbf{Missing Values} & \multirow{-2}{*}{\textbf{Replicates}} & \multirow{-2}{*}{\textbf{Sample Size}} \\ \midrule
 & & 0.05 & 10 & 100, 200, 300, 400, 500, 1000, 1500, 2000 \\ \cmidrule{3-5}
 & & 0.1 & 10 & 100, 200, 300, 400, 500, 1000, 1500, 2000 \\ \cmidrule{3-5}
\multirow{-3}{*}{\textbf{Asia}} & \multirow{-3}{*}{\begin{tabular}[c]{@{}c@{}}Random patterns\\ MNAR e 
		 MCAR\end{tabular}} & 0.2 & 10 & 100, 200, 300, 400, 500, 1000, 1500, 2000 \\ \hline
 & & 0.05 & 10 & 100, 200, 400, 800, 1200, 1600, 5000 \\ \cline{3-5}
 & \multirow{-2}{*}{\begin{tabular}[c]{@{}c@{}}Random patterns\\ MNAR e MCAR\end{tabular}} & 0.1 & 10 & 100, 200, 400, 800, 1200, 1600 \\ \cmidrule{2-5}
 &  & 0.05 & 10 & 100, 200, 400, 800, 1200, 1600, 2000 \\ \cmidrule{3-5}
\multirow{-4}{*}{\textbf{Sports}} & \multirow{-2}{*}{Most central nodes} & 0.1 & 10 & 100, 200, 400, 800, 1200, 1600 \\ \midrule
 &  & 0.01 & 8 & 200, 400, 600, 1000, 1500 \\ \cmidrule{3-5}
 & \multirow{-2}{*}{\begin{tabular}[c]{@{}c@{}}Random patterns\\ MNAR e MCAR\end{tabular}} & 0.05 & 8 & 200, 400, 600, 1000, 1500 \\ \cline{2-5}
 &  & 0.01 & 8 & 200, 400, 600, 1000, 1500 \\ \cline{3-5}
\multirow{-4}{*}{\textbf{Alarm}} & \multirow{-2}{*}{Most central nodes} & 0.05 & 8 & 200, 400, 600, 1000, 1500 \\ \hline
 & & 0.01 & 8 & 200, 400, 800, 1100 \\ \cline{3-5}
 & \multirow{-2}{*}{\begin{tabular}[c]{@{}c@{}}Random patterns \\ MNAR e MCAR\end{tabular}} & 0.05 & 8 & 400, 800, 1100 \\ \cmidrule{2-5}
 & Most central nodes & 0.01 & 8 & 200, 400, 800, 1100 \\ \cmidrule{2-5}
\multirow{-4}{*}{\textbf{Property}} & Leaves & 0.01 & 8 & 200, 400, 800, 1100 \\ \midrule
 & & 0.005 & 8 & 300, 600, 1000, 1400 \\ \cmidrule{3-5}
 & \multirow{-2}{*}{\begin{tabular}[c]{@{}c@{}}Random patterns\\ MNAR\end{tabular}} & 0.01 & 8 & 300, 600, 1000, 1400 \\ \cline{2-5}
 & Roots & 0.003 & 8 & 300, 600, 1000, 1400 \\ \cline{2-5}
 & With high degree & 0.003 & 8 & 300, 600, 1000, 1400 \\ \cline{2-5}
 & Leaves & 0.006 & 8 & 300, 600, 1000, 1400 \\ \cline{2-5}
 & \begin{tabular}[c]{@{}c@{}}Random patterns\\ MCAR\end{tabular} & 0.006 & 8 & 300, 600, 1000, 1400 \\

\multirow{-7}{*}{\textbf{ForMed}} & Most central nodes & 0.006 & 8 & 300, 600, 1000, 1400 \\ \midrule
 & & 0.005 & 8 & 300, 600, 1000, 1400 \\ \cmidrule{3-5}
 & \multirow{-2}{*}{\begin{tabular}[c]{@{}c@{}}Random patterns\\ MNAR\end{tabular}} & 0.01 & 8 & 1000 \\ \cline{2-5}
 & Most central nodes & 0.005 & 8 & 300,600,1000, 1400 \\ \cline{2-5}
 & With high degree & 0.005 & 8 & 300,600,1000 \\ \cline{2-5}
 & leaves & 0.005 & 8 & 300, 600, 1000 \\ \cline{2-5}
\multirow{-7}{*}{\textbf{Pathfinder}} & \begin{tabular}[c]{@{}c@{}}Random patterns\\ MCAR\end{tabular} & 0.005 & 8 & 300,600,1000 \\ \midrule
 & & 0.03 & 8 & 300, 600, 900, 1200 \\ \cmidrule{3-5}
 & \multirow{-2}{*}{\begin{tabular}[c]{@{}c@{}}Random patterns\\ MNAR\end{tabular}} & 0.005 & 8 & 300, 600, 900, 1200 \\ \cline{2-5}
 & \begin{tabular}[c]{@{}c@{}}Random patterns\\ MCAR\end{tabular} & 0.005 & 8 & 300, 600, 900, 1200 \\ \cmidrule{2-5}
 & Most central nodes & 0.005 & 8 & 300, 600, 900, 1200 \\ \cmidrule{2-5}
\multirow{-5}{*}{\textbf{Hailfinder}} & Leaves & 0.005 & 8 & 300, 600, 900, 1200 \\ \bottomrule
\end{tabular}%
\label{tab:list-experiments}
\end{table}

\reftitle{References}

\end{document}